\pdfoutput=1
\documentclass[dvipsnames,format=sigconf,anonymous=false,review=false]{acmart}

\usepackage{amsmath}
\usepackage{wrapfig}
\usepackage{siunitx}
\usepackage{cleveref}
\usepackage{graphicx}
\usepackage{subcaption}
\usepackage{enumitem}
\usepackage{listings}
\usepackage{float}
\usepackage{hyperref}

\AtBeginDocument{%
  \providecommand\BibTeX{{%
    \normalfont B\kern-0.5em{\scshape i\kern-0.25em b}\kern-0.8em\TeX}}}

\copyrightyear{2024}
\acmYear{2024}
\setcopyright{rightsretained}
\acmConference[GECCO '24 Companion]{Genetic and Evolutionary Computation Conference}{July 14--18, 2024}{Melbourne, VIC, Australia}
\acmBooktitle{Genetic and Evolutionary Computation Conference (GECCO '24 Companion), July 14--18, 2024, Melbourne, VIC, Australia}
\acmDOI{10.1145/3638530.3654389}
\acmISBN{979-8-4007-0495-6/24/07}

\begin{document}

\title{A Multi-objective Optimization Benchmark Test Suite for Real-time Semantic Segmentation}

\author{Yifan Zhao}
\orcid{0009-0000-8943-9500}
\affiliation{%
  \institution{Southern University of Science and Technology}
  \city{Shenzhen}
  \state{Guangdong}
  \country{China}
}
\email{AlpAcA0072@gmail.com}

\author{Zhenyu Liang}
\orcid{0009-0003-8876-8638}
\affiliation{%
  \institution{Southern University of Science and Technology}
  \city{Shenzhen}
  \state{Guangdong}
  \country{China}
}
\email{zhenyuliang97@gmail.com}

\author{Zhichao Lu}
\orcid{0000-0002-4618-3573}
\affiliation{%
  \institution{City University of Hong Kong}
  \city{Hong Kong}
  \country{China}
  \postcode{43017-6221}
}
\email{luzhichaocn@gmail.com}

\author{Ran Cheng}
\orcid{0000-0001-9410-8263}
\authornote{Corresponding Author}
\affiliation{%
  \institution{Southern University of Science and Technology}
  \city{Shenzhen}
  \state{Guangdong}
  \country{China}
}
\email{ranchengcn@gmail.com}

\renewcommand{\shortauthors}{Yifan Zhao, Zhenyu Liang, Zhichao Lu, and Ran Cheng}

\begin{abstract}
As one of the emerging challenges in Automated Machine Learning, the Hardware-aware Neural Architecture Search (HW-NAS) tasks can be treated as black-box multi-objective optimization problems (MOPs). 
An important application of HW-NAS is real-time semantic segmentation, which plays a pivotal role in autonomous driving scenarios. 
The HW-NAS for real-time semantic segmentation inherently needs to balance multiple optimization objectives, including model accuracy, inference speed, and hardware-specific considerations. 
Despite its importance, benchmarks have yet to be developed to frame such a challenging task as multi-objective optimization.
To bridge the gap, we introduce a tailored streamline to transform the task of HW-NAS for real-time semantic segmentation into standard MOPs. 
Building upon the streamline, we present a benchmark test suite, CitySeg/MOP, comprising fifteen MOPs derived from the Cityscapes dataset. 
The CitySeg/MOP test suite is integrated into the EvoXBench platform to provide seamless interfaces with various programming languages (e.g., Python and MATLAB) for instant fitness evaluations.
We comprehensively assessed the CitySeg/MOP test suite on various multi-objective evolutionary algorithms, showcasing its versatility and practicality. 
Source codes are available at \url{https://github.com/EMI-Group/evoxbench}.
\end{abstract}





\maketitle

\section{Introduction}
Neural Architecture Search (NAS), a pivotal component of Automated Machine Learning \cite{DBLP:journals/jmlr/ElskenMH19}, has traditionally focused on single-objective optimization problems, aiming to minimize prediction errors. This paradigm involves three key processes: search space design, search strategy formulation, and architecture performance evaluation.

Recent advancements have introduced NAS with multi-objective optimization, particularly in Hardware-aware Neural Architecture Search (HW-NAS). 
HW-NAS automates the search for deep neural network architectures, aligning them with specific applications \cite{DBLP:conf/iclr/LiYFZZYY0HL21}, which is crucial for resource-constrained platforms. 
Real-time semantic segmentation emerges as a key challenge in this domain, meeting the need for high computational efficiency with accurate real-time processing.

Semantic segmentation, a cornerstone of computer vision, assigns semantic labels to each image pixel. However, the shift towards real-time applications, such as autonomous driving, has underscored the necessity of considering multiple objectives in model design, including not only hardware-unrelated metrics such as model accuracy (measured as mean Intersection over Union, $mIoU$), the scale of models but also hardware-related inference speed.
This emerging demand highlights the significance of HW-NAS in devising architectures that successfully balance these diverse optimization objectives.

Due to the multiple objectives in designing and optimizing the DNN architectures, the HW-NAS tasks can be fundamentally viewed as a Multi-objective Optimization Problem (MOP).
Meanwhile, since the black-box optimization characteristics of the HW-NAS, conventional optimization methods encounter significant challenges in addressing such problems. 

Therefore, it is intuitive to adopt representative Multi-Objective Evolutionary Algorithms (MOEAs) \cite{10004638} for HW-NAS. 
However, the application of MOEAs within the HW-NAS field is relatively under-explored. 
Existing NAS work on semantic segmentation\cite{DBLP:conf/cvpr/LiuCSAHY019} lacks capabilities of hardware awareness or constraints on real-time.

To address the challenges, this paper proposes a multi-objective optimization benchmark test suite for real-time semantic segmentation, dubbed CitySeg/MOP. The benchmark introduces an architecture search space for real-time semantic segmentation and tailors two categories of performance evaluators covering the multiple objectives. 
The benchmark test suite is seamlessly integrated into the platform EvoXBench \cite{10004638} and provides interfaces with various programming languages (e.g., Python and MATLAB) without the requirement for any machine learning accelerating libraries. This fills the gap left by the lack of a multi-objective optimization benchmark test suite for real-time semantic segmentation in the HW-NAS.

\section{Method}

\begin{figure}
  \centering
  \includegraphics[width=0.98\linewidth]{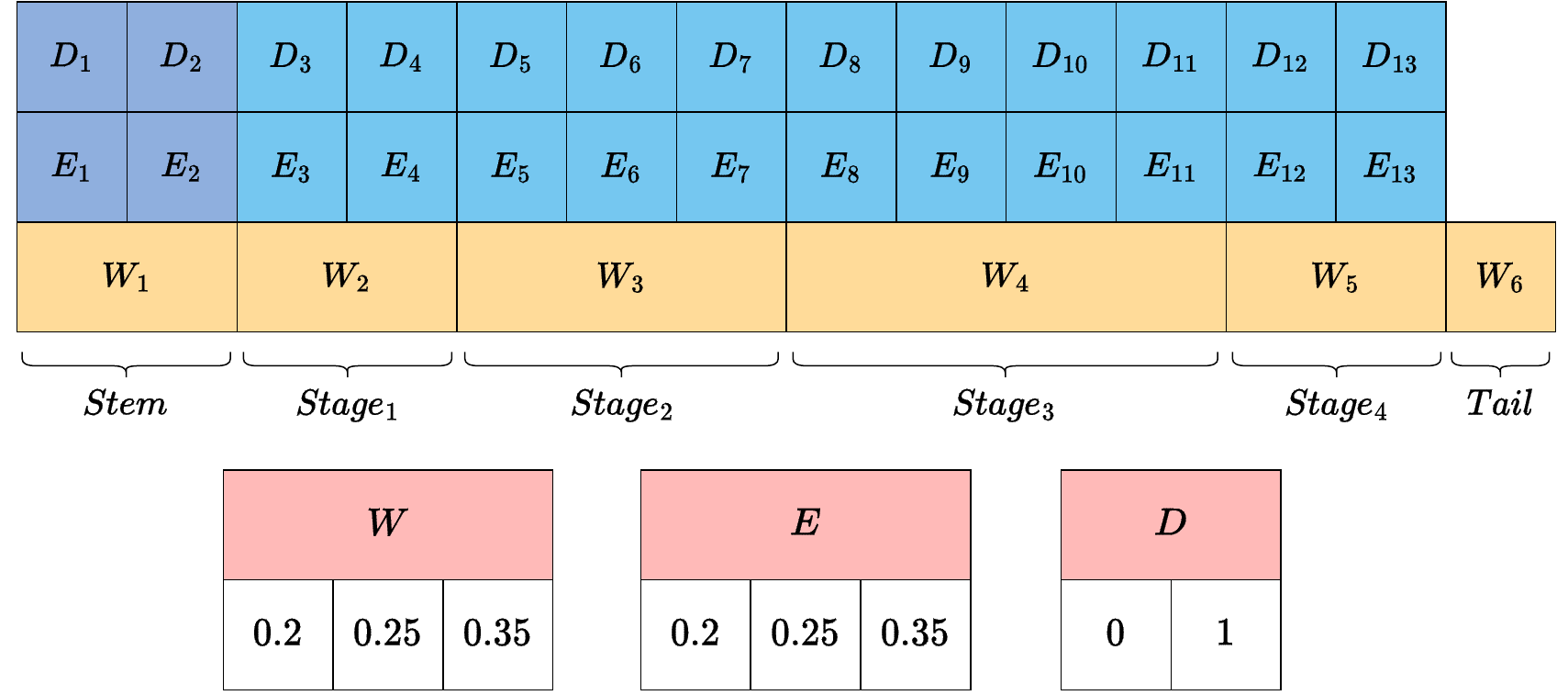}
  \caption{Architecture encoding. The search space is encoded as a 32-bit fixed-length integer-valued string.}
  \label{fig:specific search space}
\end{figure}

\subsection{Problem Formulation}
The HW-NAS problem can be formulated as a MOP, which can be defined as follows \cite{10004638}:
\begin{equation}
  \label{eq: HW-NAS}
  \begin{aligned}
    \min_{\boldsymbol{x}}&\ \boldsymbol{F}(\boldsymbol{x})~=~\left(f^{e}(\boldsymbol{x};\boldsymbol{\omega}^{*}(\boldsymbol{x})),\ \boldsymbol{f}^{c}(\boldsymbol{x}),\ \boldsymbol{f}^{\mathcal{H}}(\boldsymbol{x})\right)\\
    \mathrm{s.t.}\ &\boldsymbol{\omega}^{*}(\boldsymbol{x})=\underset{\boldsymbol{\omega}}{\mathrm{argmin}}\left.\mathcal{L}(\boldsymbol{x};\boldsymbol{\omega}),\quad\boldsymbol{x}\in\Omega_{\boldsymbol{x}}\right.
  \end{aligned},
\end{equation}
where
\begin{equation}
  \label{eq: fc}
  \boldsymbol{f}^c(\boldsymbol{x})~=~
  \begin{pmatrix}
    f_{1}(\boldsymbol{x}) & \cdots & f_{m}(\boldsymbol{x})
  \end{pmatrix},
\end{equation}
and 
\begin{equation}
  \label{eq: fh}
  \begin{aligned}
  \boldsymbol{f}^{\mathcal{H}}(\boldsymbol{x})&~=~
  \begin{pmatrix}
    f_{1}^{h_1}(\boldsymbol{x}) & \cdots & f_{n}^{h_1}(\boldsymbol{x}) \\
    \vdots & \ddots & \vdots \\
    f_{1}^{h_{|\mathcal{H}|}}(\boldsymbol{x}) & \cdots & f_{n}^{h_{|\mathcal{H}|}}(\boldsymbol{x}) \\
  \end{pmatrix},\\
  \mathcal{H}&~=~\{h_1,\ldots,h_{|\mathcal{H}|}\},
  \end{aligned}
\end{equation}
where $\Omega_{\boldsymbol{x}}$ is the architecture search space, $\boldsymbol{x}$ is an encoded network architecture, $\boldsymbol{\omega}(\boldsymbol{x})$ is the weights vector of the network architecture $\boldsymbol{x}$. 
For a dataset $\mathcal{D}$, there is $\mathcal{D}=\{\mathcal{D}_{\text{train}},\mathcal{D}_{\text{valid}},\mathcal{D}_{\text{test}}\}$. $\boldsymbol{\omega}^{*}(\boldsymbol{x})$ is the optimal weights vector of the network architecture $\boldsymbol{x}$ that minimizes loss $\mathcal{L}$ on $\mathcal{D}_{\text{valid}}$. 
$\boldsymbol{F}(\boldsymbol{x})$ is the objective vector, which consists of three categories: prediction error ($f^{e}$), complexity-related objectives ($\boldsymbol{f}^{c}$), and hardware-related objectives ($\boldsymbol{f}^{\mathcal{H}}$), in which $\mathcal{H}$ denotes the set of hardware platforms. 
In $f_{n}^{h_{|\mathcal{H}|}}(\boldsymbol{x})$, the $h_{|\mathcal{H}|}$ denotes the ${|\mathcal{H}|}$-th hardware platform and $n$ denotes the ${n}$-th objective.



\subsection{Search Space}

The search space adopted is inspired by MoSegNAS \cite{DBLP:journals/tai/LuCHZQY23}. The search space's overall structure comprises one stem layer, four stage layers, and one tail layer. The stem and stage layers contain cells that need to be searched, while the tail layer remains constant to be a global average pooling layer.
Each cell comprises a sequence of convolutions, which is a typical residual bottleneck block.

In addition, the search space encodes essential attributes, including depth, expansion ratio, and width (denoting the channels). 
The specific form of the coded architecture is shown in Figure~\ref{fig:specific search space}.

\subsection{Fitness Evaluator}
We use different fitness evaluators to cope with the diversity of the objectives. There are two categories of fitness evaluators and three categories of objectives, where the surrogate model is used for $f^e$, and the look-up table is used for $\boldsymbol{f}^c$ and $\boldsymbol{f}^{\mathcal{H}}$.



The fitness evaluator of inference accuracy objective ($f^e$) is measured in $(1 - mIoU)$. The $mIoU$ is defined as follows:
\begin{equation}
  \label{eq: miou}
  mIoU=\frac1N\sum_{n=1}^NIoU=\frac1N\sum_{n=1}^N\frac{TP}{(TP+FP+FN)},
\end{equation}
where $TP$, $FP$, and $FN$ are the numbers of true positive, false positive, and false negative pixels of the segmentation result in the given image, and $N$ is the number of the defined classes over the whole test set.

It is computationally unacceptable to train from scratch to assess the prediction accuracy. Rather than adopting the proxy methods (with fewer training epochs or gradient approximation\cite{DBLP:conf/iclr/BakerGRN18}), we adopted the surrogate model to approximate the mapping from the network architectures' space to the inference accuracy objective ($f^e$). 
In this paper, we adopt a Multi-Layer Perceptron (MLP) with ranking loss and mean square error (MSE) loss as the surrogate model:
\begin{equation}
  \begin{alignedat}{1}
    \mathcal{L}& = \mathcal{L}_{\text{rank}} + \mathcal{L}_{\text{mse}},\\
    \mathcal{L}_{\text{mse}}& = \frac1N\sum_i\|\hat{y}_i-y_i\|^2,\\
    \mathcal{L}_{\text{rank}}& = \frac{1}{2N}\sum_{i,j}\text{max}\big(0,\gamma-\delta(\hat{y}_i,\hat{y}_j)(y_i-y_j)\big), \\
    \delta(\hat{y}_i,\hat{y}_j)& = \begin{cases}1,\text{if }\hat{y}_i>\hat{y}_j,\\-1,\text{otherwise,}\end{cases}
  \end{alignedat}
\end{equation}
where the $\hat{y}_i$ is the output of the MLP when given input $x_i$. 
The $\gamma$ is the margin controller hyperparameter set to $5 \times 10^{-2}$. 

\begin{figure}
  \centering
  \includegraphics[width=0.95\linewidth]{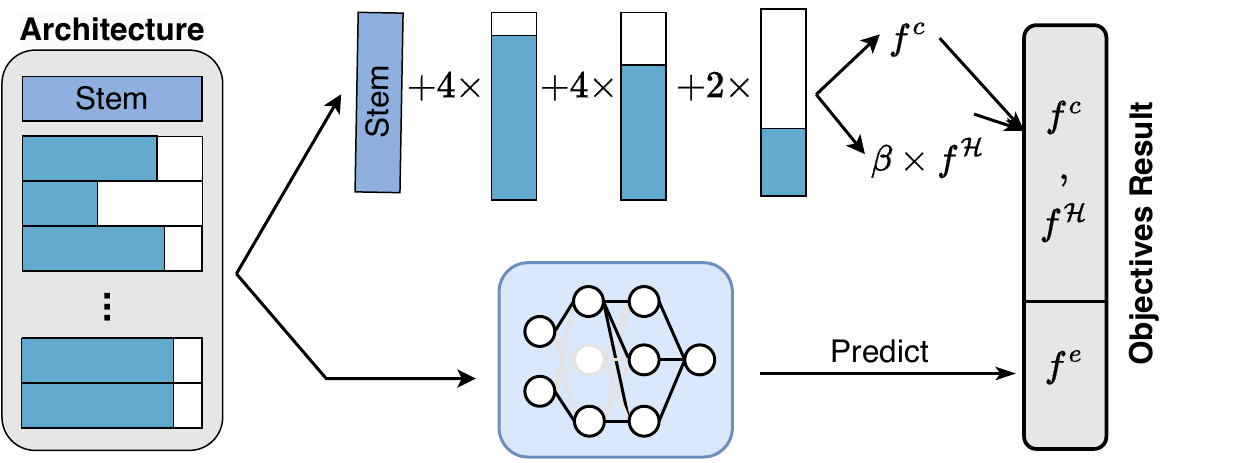}
  \caption{An example process of performance evaluation via fitness evaluator. The upper half and the bottom half indicate an example process of the look-up predictor and the surrogate model predictor respectively.}
  \label{fig:fitness evaluator}
\end{figure}

We have the assumption that complexity-related objectives ($\boldsymbol{f}^c$) and the hardware-related objectives ($\boldsymbol{f}^{\mathcal{H}}$) are decomposable, i.e.~the objectives can be decomposed to the sum up of the layers' corresponding objectives. Under this assumption, we can enumerate all the values of the stems and the stages individually to gain the look-up table. Thus, we can compose the architectures' stem and stages by summing up the values of $\boldsymbol{f}^c$ and $\boldsymbol{f}^{\mathcal{H}}$ to gain the overall result.
The procedure of the whole fitness evaluation is shown in Figure~\ref{fig:fitness evaluator}. 
The $\boldsymbol{f}^\mathcal{H}$ can be affected by stochastic perturbation with the coefficient $\beta$ to simulate the hardware's performance fluctuation. The $\beta$ is set to $[0.95, 1.05]$ for $f^\mathcal{H}_1$ and $[0.98, 1.02]$ for $f^\mathcal{H}_2$ in this paper.

Unlike the stochastic gradient descent process, the approximated optimal weights of the candidate network architecture are inherited from the trained hypernetwork. 
Hypernetwork essentially introduces a mapping from the network architecture space $\mathbb{Z}^d$ to the weight space $\mathbb {R}^m$, i.e.~$\Psi:\ \mathbb{Z}^d \rightarrow \mathbb {R}^m$. $\Psi$ can be formulated as follows:
\begin{equation}\label{eq:psi}
  \begin{gathered}
      \Psi(\boldsymbol{x}) = \underset{\boldsymbol{\omega}}{{\rm argmin}}\ \mathcal{L}(\boldsymbol{x};\boldsymbol{\omega}),
  \end{gathered}
\end{equation}
where $d$ is the dimension of network architecture space $\mathbb{Z}$ and $m$ is the dimension of weights space $\mathbb{R}$.

\begin{table}[ht]
  \caption{Definition of the CitySeg/MOP test suite.}
  \label{tab: definition of CitySeg/MOP}
  \begin{tabular*}{\hsize}{@{}@{\extracolsep{\fill}}ccccc@{}}
  \toprule
  Problem  &$D$  &$M$  &Objectives \\
  \midrule
  \rule{0pt}{10pt} CitySeg/MOP1   &32  &2  &$f^e, f^{h_1}_1$ \\
  \rule{0pt}{10pt} CitySeg/MOP2  &32   &3   &$f^e, f^{h_1}_1, f^c_1$ \\
  \rule{0pt}{10pt} CitySeg/MOP3 &32 &3 &$f^e, f^{h_1}_1, f^c_2$ \\
  \rule{0pt}{10pt} CitySeg/MOP4  &32  &4  &$f^e, f^{h_1}_1, f^{h_1}_2, f^c_1$ \\
  \rule{0pt}{10pt} CitySeg/MOP5  &32  &5  &$f^e, f^{h_1}_1, f^{h_1}_2, f^c_1, f^c_2$ \\ 
  \noalign{\smallskip}\hline\noalign{\smallskip}
  
  \rule{0pt}{10pt} CitySeg/MOP6  &32  &2  &$f^e, f^{h_2}_1$ \\
  \rule{0pt}{10pt} CitySeg/MOP7  &32  &3  &$f^e, f^{h_2}_1, f^c_1$ \\
  \rule{0pt}{10pt} CitySeg/MOP8  &32  &3  &$f^e, f^{h_2}_1, f^c_2$ \\
  \rule{0pt}{10pt} CitySeg/MOP9  &32  &4  &$f^e, f^{h_2}_1, f^{h_2}_2, f^c_1$ \\
  \rule{0pt}{10pt} CitySeg/MOP10  &32  &5  &$f^e, f^{h_2}_1, f^{h_2}_2, f^c_1, f^c_2$ \\
  \noalign{\smallskip}\hline\noalign{\smallskip}
  \rule{0pt}{10pt} CitySeg/MOP11  &32  &3  &$f^e, f^{h_1}_1, f^{h_2}_1$ \\
  \rule{0pt}{10pt} CitySeg/MOP12  &32  &5  &$f^e, f^{h_1}_1, f^{h_2}_1, f^{h_1}_2, f^{h_2}_2$ \\
  \rule{0pt}{10pt} CitySeg/MOP13  &32  &6  &$f^e, f^{h_1}_1, f^{h_2}_1, f^{h_1}_2, f^{h_2}_2, f^c_1$ \\
  \rule{0pt}{10pt} CitySeg/MOP14  &32  &6  &$f^e, f^{h_1}_1, f^{h_2}_1, f^{h_1}_2, f^{h_2}_2, f^c_2$ \\
  \rule{0pt}{10pt} CitySeg/MOP15  &32  &7  &$f^e, f^{h_1}_1, f^{h_2}_1, f^{h_1}_2, f^{h_2}_2, f^c_1, f^c_2$ \\

\bottomrule
\end{tabular*}
\end{table}

\subsection{Benchmark Generation}
The network architectures are trained and tested on the Cityscapes dataset \cite{DBLP:conf/cvpr/CordtsORREBFRS16}, which is a large-scale dataset for urban city street semantic segmentation. 
Input images with a low expand ratio can significantly reduce memory usage and computation resource overhead.
Thus, we provide images at different expansion ratios in the search space.

\begin{figure}
  \centering
  \newcommand{\subfigwidth}{0.32\linewidth} 
  \begin{subfigure}{\subfigwidth}
    \centering
    \includegraphics[width=\linewidth]{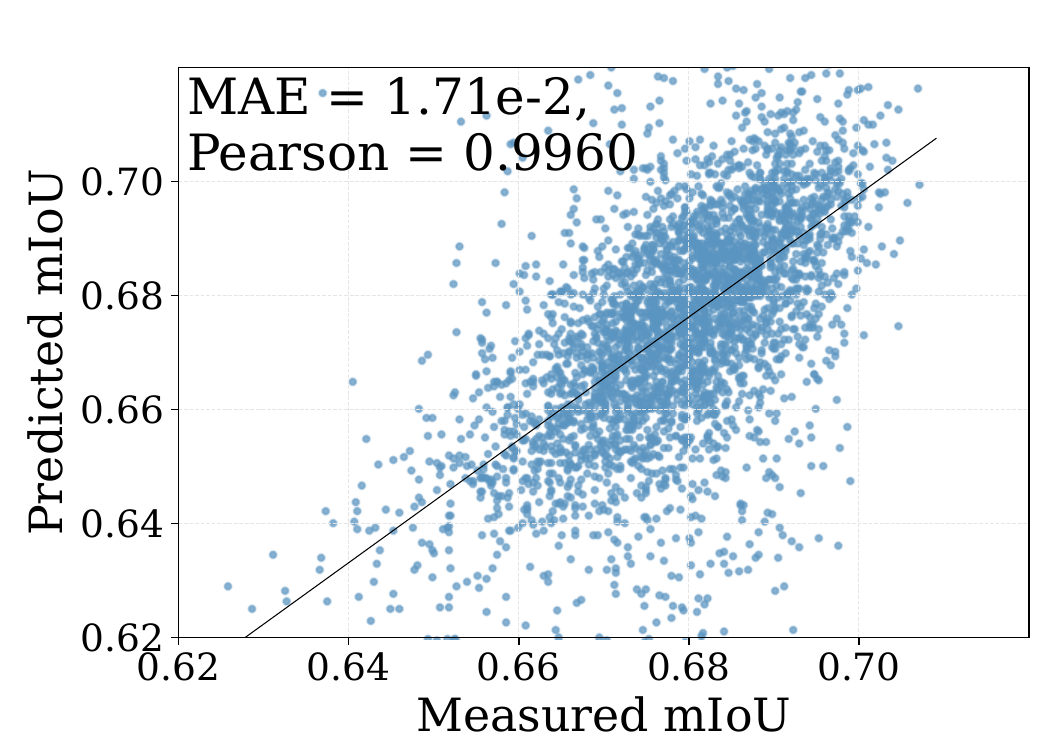}
    \caption{$f^e$}
    \label{fig:miou corrolation}
  \end{subfigure}
  \hfill
  \begin{subfigure}{\subfigwidth}
    \centering
    \includegraphics[width=\linewidth]{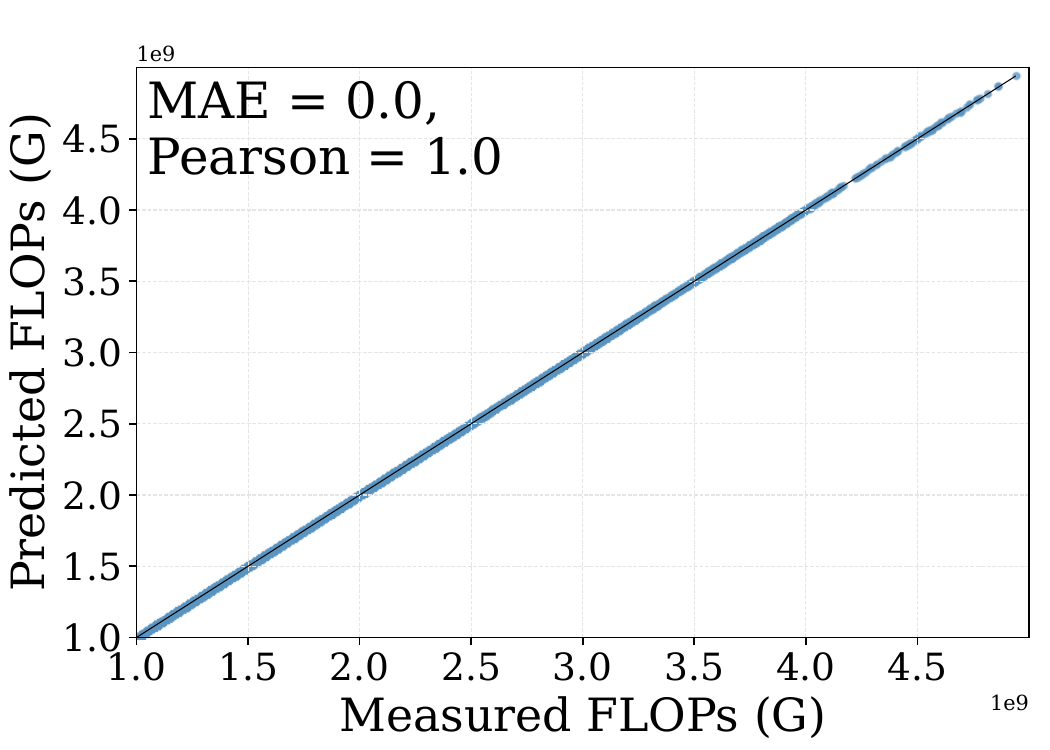}
    \caption{$f^c_1$}
    \label{fig:flops corrolation}
  \end{subfigure}
  \hfill
  \begin{subfigure}{\subfigwidth}
    \centering
    \includegraphics[width=\linewidth]{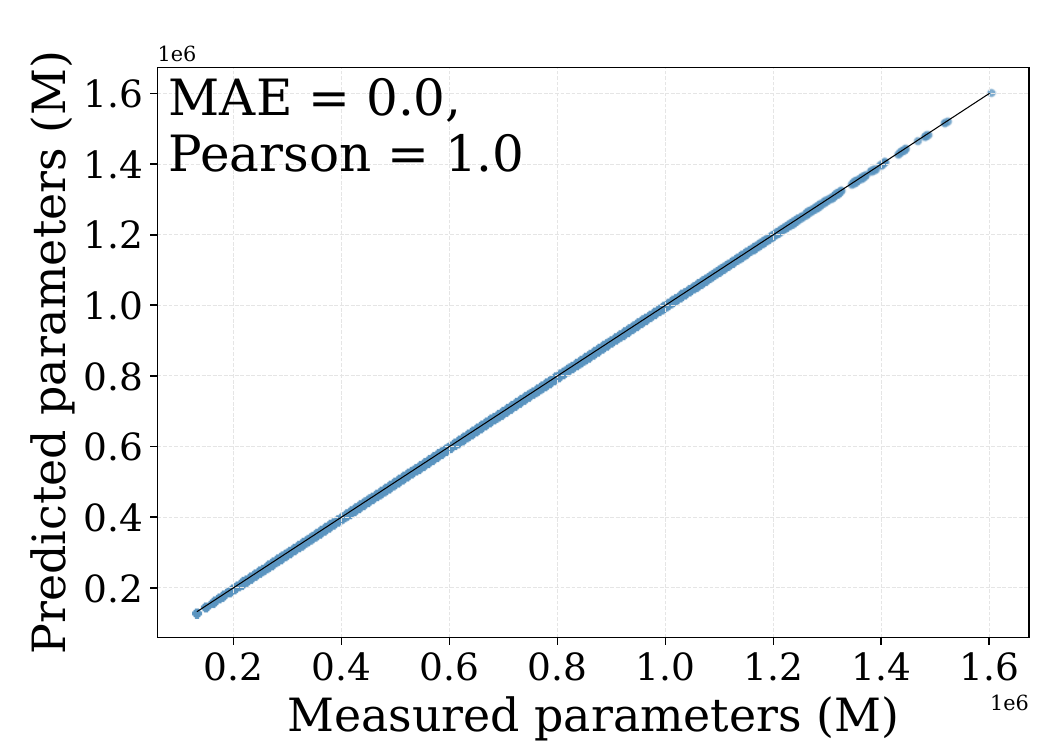}
    \caption{$f^c_2$}
    \label{fig:params corrolation}
  \end{subfigure}
  \caption{Correlations between the measured and the predicted metrics. 
  The Mean Absolute Error (MAE) and Pearson correlation coefficient ($\rho$) are shown in the corresponding subfigures.} 
  \label{fig:corrolation}
\end{figure}

Based on EvoXBench \cite{10004638}, we construct a tailored benchmark test suite CitySeg/MOP for real-time semantic segmentation. As summarized in Table~\ref{tab: definition of CitySeg/MOP}, there are fifteen instances in total, $D$ indicates the number (dimension) of decision variables, and $M$ indicates the number of objectives in the problems. 
For deployment, there are two hardware devices involved: the NVIDIA GeForce RTX 4090 ($h_1$) and Huawei Atlas 200 DK ($h_2$), where the former is a representative GPU device and the latter is a typical ARM-based edge computing device with only 9.5W energy consumption. 
The CitySeg/MOP test suite's problems are divided into three categories. The first five problems contain $h_1$'s objectives, and the sixth to the tenth problems contain $h_2$'s objectives. The last five problems enable the hardware-aware features by combining $h_1$'s and $h_2$'s objectives.
Table~\ref{table:definition of objectives} in Appendix \ref{appendix:Experimental Setup} provides the definition of objectives in CitySeg/MOP test suite.

As summarized in Table~\ref{tab: definition of CitySeg/MOP}, the test instances are listed for hardware and number of objectives in ascending order, from two to seven objectives. 

\section{Experiments}
This section provides the experimental results for comprehensively assessing the proposed benchmark test suite CitySeg/MOP. 
Firstly, we conducted experiments to assess the prediction accuracy, sample diversity, and evaluation efficiency of the predictors; afterward, we conducted benchmark tests running on the proposed test suite.
The results of evaluation efficiency and benchmark tests are obtained by running each algorithm 31 times on EvoX \cite{evox} or PlatEMO \cite{PlatEMO} independently, with 10,000 fitness evaluations. 


\subsection{Prediction Accuracy}

The correlations between the measured and the predicted metrics are shown in Figure~\ref{fig:corrolation}. For Figure~\ref{fig:flops corrolation} and Figure~\ref{fig:params corrolation}, we measured the number of floating point operations (FLOPs) ($f^c_1$) and the number of parameters ($f^c_2$) of the smallest unit. Network architectures' FLOPs ($f^c_1$) and parameters ($f^c_2$) are weighed against the corresponding parameters of the smallest units. The Mean Absolute Error (MAE) of the FLOPs and the parameters is $0.0$. For Figure~\ref{fig:miou corrolation}, the MAE and $\rho$ of the prediction error ($f^e$) are $1.71\times 10^{-2}$ and $0.9960$ respectively. Figure~\ref{fig:corrolation} shows the accuracy of the predictions for $f^e$, $f^c_1$ and $f^c_2$.

\subsection{Sample Diversity}

\begin{figure}
  \centering
  \newcommand{\subfigwidth}{0.48\linewidth}
  \begin{subfigure}{\subfigwidth}
    \centering
    \includegraphics[width=\linewidth]{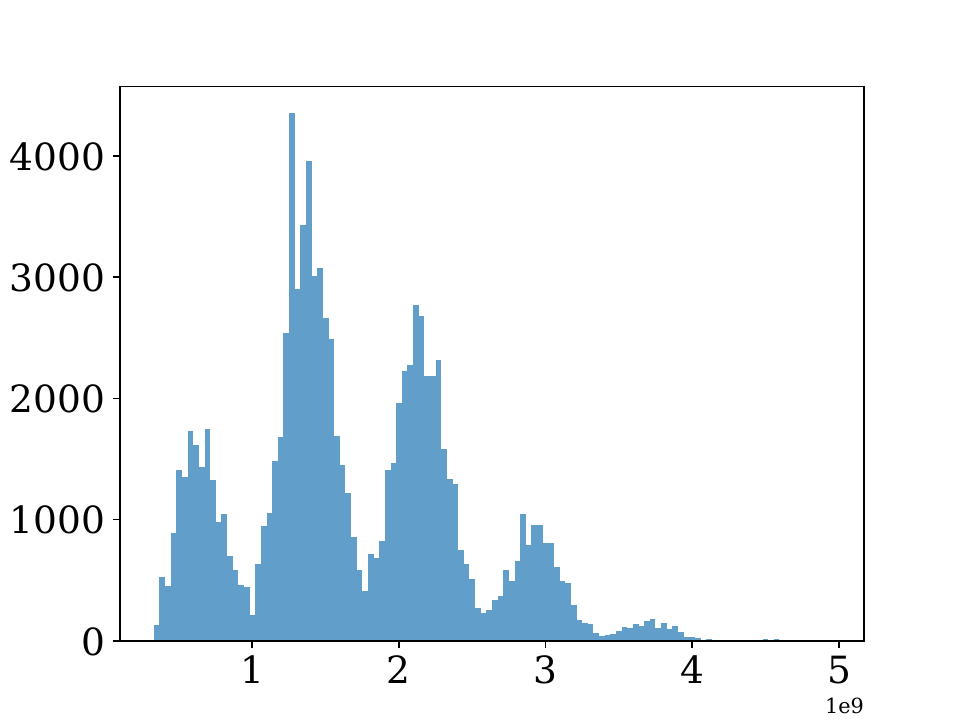}
    \caption{$f^c_1$'s distribution.}
    \label{fig:flops distribution}
  \end{subfigure}
  \hfill
  \begin{subfigure}{\subfigwidth}
    \centering
    \includegraphics[width=\linewidth]{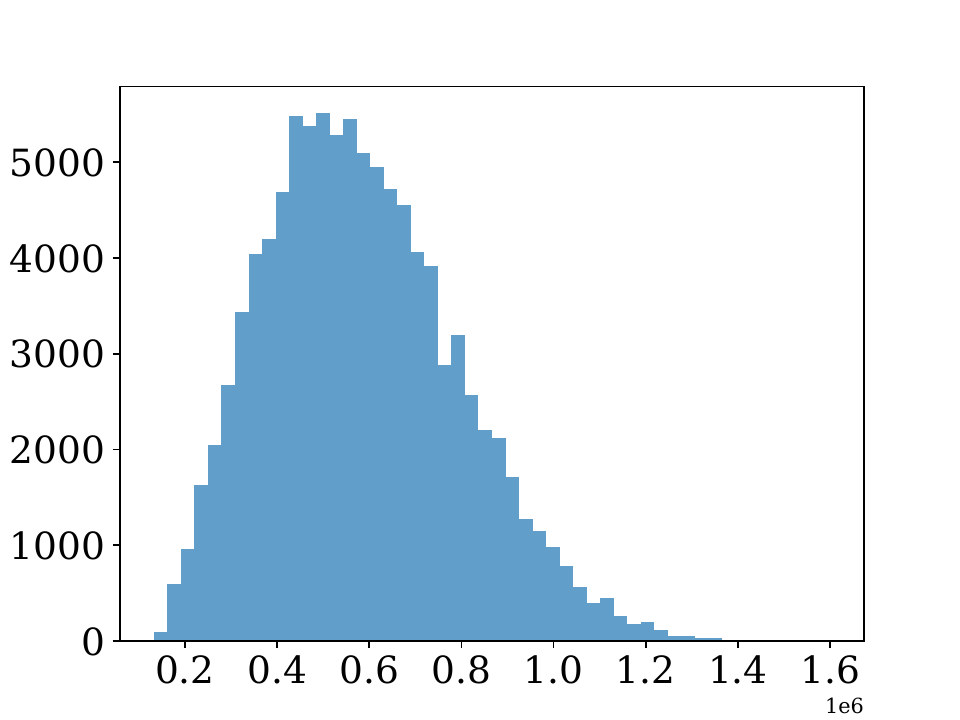}
    \caption{$f^c_2$'s distribution.}
    \label{fig:params distribution}
  \end{subfigure}
  \caption{Distribution of $f^c_1$ and $f^c_2$ under randomly sampled architectures.}
  
  \label{fig:random sampled architectures histogram}
\end{figure}

Figure~\ref{fig:random sampled architectures histogram} shows the randomly sampled architectures' distribution of FLOPs ($f^c_1$) and number of parameters ($f^c_2$). 
The FLOPs ($f^c_1$) has multiple peaks, which is normally distributed, and the values within each peak are also normally distributed. 
The short tail on the left occurs because network architectures with the depth of all stages being zero are not permitted. Conversely, the long-tailed peaks on the right represent scenarios where no stage has zero depth.

Figure~\ref{fig:architectures' correlation between consumption and flops and params} in Appendix \ref{appendix:Supplementary Results} shows the randomly sampled architectures' correlation between hardware-related consumption and FLOPs, the number of parameters. We can see from Figure~\ref{fig:$f^c_1$ and $f^h_2$'s correlation.} that data is distributed in multiple clusters, resulting from the stages in the search space. This is consistent with the data distribution in Figure~\ref{fig:flops distribution}.

\subsection{Evaluation Efficiency}
Table~\ref{table:time cost} in Appendix \ref{appendix:Supplementary Results} provides the evaluation efficiency when called by Python or MATLAB's interface. 
The evaluation is performed on a local machine with one CPU core using PlatEMO \cite{PlatEMO} for MATLAB, and NVIDIA GeForce RTX 4090 using EvoX \cite{evox} for Python respectively.

\subsection{Benchmark Tests}

As a demonstration, we conducted benchmark tests on the proposed CitySeg/MOP test suite using six representative MOEAs falling into three different categories: (1)  NSGA-II \cite{DBLP:journals/tec/DebAPM02} and NSGA-III \cite{DBLP:journals/tec/DebJ14} (dominance-based); (2) MOEA/D \cite{DBLP:journals/tec/ZhangL07} and RVEA\footnote{We use the version with the reference vector regeneration strategy, which is also known as the RVEA*.} \cite{7386636} (decomposition based); (3) IBEA \cite{DBLP:conf/ppsn/ZitzlerK04} and HypE \cite{DBLP:journals/ec/BaderZ11} (indicator based).

Table~\ref{tab: problem result of CitySeg/MOP} in Appendix \ref{appendix:Supplementary Results} summarizes the statistical results of the HV values of the test instances on the CitySeg/MOP test suite. 
Of these test instances, in terms of solving test instances with MOPs and MaOPs (Problems with more than three objectives), algorithms NSGA-II, NSGA-III (dominance-based methods) have an advantage. IBEA, HypE (indicator based methods) perform better on certain problems.

Figure~\ref{fig:citymop problems} in Appendix \ref{appendix:Supplementary Results} shows the non-dominated solutions obtained by algorithms.
By performing MOEAs on CitySeg/MOP15, algorithms favor low energy-consumption, low-latency network architectures. However, these algorithms are inconsistent in the choice of prediction error objective and model scale objectives.

\section{Conclusion}

In this paper, we introduce CitySeg/MOP, a multi-objective optimization benchmark test suite tailored for real-time semantic segmentation. 
By considering HW-NAS tasks real-time semantic segmentation as standard MOPs, CitySeg/MOP encompasses various objectives such as accuracy, computational efficiency, and real-time constraints. 
The experimental results demonstrate the efficiency, accuracy and diversity of CitySeg/MOP. 
Moreover, by providing seamless integration to the EvoXBench \cite{10004638} platform, it enables efficient interfaces to general MOEAs implemented in various programming languages.
As a standard multi-objective optimization benchmark test suite tailored for real-time semantic segmentation, CitySeg/MOP contributes to filling the gap between training EMO research and its applications in complex HW-NAS tasks.



\bibliographystyle{ACM-Reference-Format}
\bibliography{sample-base}


\begin{thebibliography}{15}


\ifx \showCODEN    \undefined \def \showCODEN     #1{\unskip}     \fi
\ifx \showDOI      \undefined \def \showDOI       #1{#1}\fi
\ifx \showISBNx    \undefined \def \showISBNx     #1{\unskip}     \fi
\ifx \showISBNxiii \undefined \def \showISBNxiii  #1{\unskip}     \fi
\ifx \showISSN     \undefined \def \showISSN      #1{\unskip}     \fi
\ifx \showLCCN     \undefined \def \showLCCN      #1{\unskip}     \fi
\ifx \shownote     \undefined \def \shownote      #1{#1}          \fi
\ifx \showarticletitle \undefined \def \showarticletitle #1{#1}   \fi
\ifx \showURL      \undefined \def \showURL       {\relax}        \fi
\providecommand\bibfield[2]{#2}
\providecommand\bibinfo[2]{#2}
\providecommand\natexlab[1]{#1}
\providecommand\showeprint[2][]{arXiv:#2}

\bibitem[Bader and Zitzler(2011)]%
        {DBLP:journals/ec/BaderZ11}
\bibfield{author}{\bibinfo{person}{Johannes Bader} {and} \bibinfo{person}{Eckart Zitzler}.} \bibinfo{year}{2011}\natexlab{}.
\newblock \showarticletitle{HypE: An Algorithm for Fast Hypervolume-Based Many-Objective Optimization}.
\newblock \bibinfo{journal}{\emph{IEEE Transactions on Evolutionary Computation}} \bibinfo{volume}{19}, \bibinfo{number}{1} (\bibinfo{year}{2011}), \bibinfo{pages}{45--76}.
\newblock


\bibitem[Baker et~al\mbox{.}(2018)]%
        {DBLP:conf/iclr/BakerGRN18}
\bibfield{author}{\bibinfo{person}{Bowen Baker}, \bibinfo{person}{Otkrist Gupta}, \bibinfo{person}{Ramesh Raskar}, {and} \bibinfo{person}{Nikhil Naik}.} \bibinfo{year}{2018}\natexlab{}.
\newblock \showarticletitle{Accelerating Neural Architecture Search using Performance Prediction}. In \bibinfo{booktitle}{\emph{6th International Conference on Learning Representations, {ICLR} 2018, Vancouver, BC, Canada, April 30 - May 3, 2018, Workshop Track Proceedings}}.
\newblock


\bibitem[Cheng et~al\mbox{.}(2016)]%
        {7386636}
\bibfield{author}{\bibinfo{person}{Ran Cheng}, \bibinfo{person}{Yaochu Jin}, \bibinfo{person}{Markus Olhofer}, {and} \bibinfo{person}{Bernhard Sendhoff}.} \bibinfo{year}{2016}\natexlab{}.
\newblock \showarticletitle{A Reference Vector Guided Evolutionary Algorithm for Many-Objective Optimization}.
\newblock \bibinfo{journal}{\emph{IEEE Transactions on Evolutionary Computation}} \bibinfo{volume}{20}, \bibinfo{number}{5} (\bibinfo{year}{2016}), \bibinfo{pages}{773--791}.
\newblock


\bibitem[Cordts et~al\mbox{.}(2016)]%
        {DBLP:conf/cvpr/CordtsORREBFRS16}
\bibfield{author}{\bibinfo{person}{Marius Cordts}, \bibinfo{person}{Mohamed Omran}, \bibinfo{person}{Sebastian Ramos}, \bibinfo{person}{Timo Rehfeld}, \bibinfo{person}{Markus Enzweiler}, \bibinfo{person}{Rodrigo Benenson}, \bibinfo{person}{Uwe Franke}, \bibinfo{person}{Stefan Roth}, {and} \bibinfo{person}{Bernt Schiele}.} \bibinfo{year}{2016}\natexlab{}.
\newblock \showarticletitle{The Cityscapes Dataset for Semantic Urban Scene Understanding}. In \bibinfo{booktitle}{\emph{2016 {IEEE} Conference on Computer Vision and Pattern Recognition, {CVPR} 2016, Las Vegas, NV, USA, June 27-30, 2016}}. \bibinfo{pages}{3213--3223}.
\newblock


\bibitem[Deb et~al\mbox{.}(2002)]%
        {DBLP:journals/tec/DebAPM02}
\bibfield{author}{\bibinfo{person}{Kalyanmoy Deb}, \bibinfo{person}{Samir Agrawal}, \bibinfo{person}{Amrit Pratap}, {and} \bibinfo{person}{T. Meyarivan}.} \bibinfo{year}{2002}\natexlab{}.
\newblock \showarticletitle{A Fast and Elitist Multiobjective Genetic Algorithm: {NSGA-II}}.
\newblock \bibinfo{journal}{\emph{{IEEE} Transactions on Evolutionary Computation}} \bibinfo{volume}{6}, \bibinfo{number}{2} (\bibinfo{year}{2002}), \bibinfo{pages}{182--197}.
\newblock


\bibitem[Deb and Jain(2014)]%
        {DBLP:journals/tec/DebJ14}
\bibfield{author}{\bibinfo{person}{Kalyanmoy Deb} {and} \bibinfo{person}{Himanshu Jain}.} \bibinfo{year}{2014}\natexlab{}.
\newblock \showarticletitle{An Evolutionary Many-Objective Optimization Algorithm Using Reference-Point-Based Nondominated Sorting Approach, Part {I:} Solving Problems With Box Constraints}.
\newblock \bibinfo{journal}{\emph{{IEEE} Transactions on Evolutionary Computation}} \bibinfo{volume}{18}, \bibinfo{number}{4} (\bibinfo{year}{2014}), \bibinfo{pages}{577--601}.
\newblock


\bibitem[Elsken et~al\mbox{.}(2019)]%
        {DBLP:journals/jmlr/ElskenMH19}
\bibfield{author}{\bibinfo{person}{Thomas Elsken}, \bibinfo{person}{Jan~Hendrik Metzen}, {and} \bibinfo{person}{Frank Hutter}.} \bibinfo{year}{2019}\natexlab{}.
\newblock \showarticletitle{Neural Architecture Search: {A} Survey}.
\newblock \bibinfo{journal}{\emph{Journal of Machine Learning Research}}  \bibinfo{volume}{20} (\bibinfo{year}{2019}), \bibinfo{pages}{55:1--55:21}.
\newblock


\bibitem[Huang et~al\mbox{.}(2024)]%
        {evox}
\bibfield{author}{\bibinfo{person}{Beichen Huang}, \bibinfo{person}{Ran Cheng}, \bibinfo{person}{Zhuozhao Li}, \bibinfo{person}{Yaochu Jin}, {and} \bibinfo{person}{Kay~Chen Tan}.} \bibinfo{year}{2024}\natexlab{}.
\newblock \showarticletitle{{EvoX}: {A} {Distributed} {GPU}-accelerated {Framework} for {Scalable} {Evolutionary} {Computation}}.
\newblock \bibinfo{journal}{\emph{IEEE Transactions on Evolutionary Computation}} (\bibinfo{year}{2024}).
\newblock
\urldef\tempurl%
\url{https://doi.org/10.1109/TEVC.2024.3388550}
\showDOI{\tempurl}


\bibitem[Li et~al\mbox{.}(2021)]%
        {DBLP:conf/iclr/LiYFZZYY0HL21}
\bibfield{author}{\bibinfo{person}{Chaojian Li}, \bibinfo{person}{Zhongzhi Yu}, \bibinfo{person}{Yonggan Fu}, \bibinfo{person}{Yongan Zhang}, \bibinfo{person}{Yang Zhao}, \bibinfo{person}{Haoran You}, \bibinfo{person}{Qixuan Yu}, \bibinfo{person}{Yue Wang}, \bibinfo{person}{Cong Hao}, {and} \bibinfo{person}{Yingyan Lin}.} \bibinfo{year}{2021}\natexlab{}.
\newblock \showarticletitle{HW-NAS-Bench: Hardware-Aware Neural Architecture Search Benchmark}. In \bibinfo{booktitle}{\emph{9th International Conference on Learning Representations, {ICLR} 2021, Virtual Event, Austria, May 3-7, 2021}}.
\newblock


\bibitem[Liu et~al\mbox{.}(2019)]%
        {DBLP:conf/cvpr/LiuCSAHY019}
\bibfield{author}{\bibinfo{person}{Chenxi Liu}, \bibinfo{person}{Liang{-}Chieh Chen}, \bibinfo{person}{Florian Schroff}, \bibinfo{person}{Hartwig Adam}, \bibinfo{person}{Wei Hua}, \bibinfo{person}{Alan~L. Yuille}, {and} \bibinfo{person}{Li Fei{-}Fei}.} \bibinfo{year}{2019}\natexlab{}.
\newblock \showarticletitle{Auto-DeepLab: Hierarchical Neural Architecture Search for Semantic Image Segmentation}. In \bibinfo{booktitle}{\emph{{IEEE} Conference on Computer Vision and Pattern Recognition, {CVPR} 2019, Long Beach, CA, USA, June 16-20, 2019}}. \bibinfo{pages}{82--92}.
\newblock


\bibitem[Lu et~al\mbox{.}(2023)]%
        {DBLP:journals/tai/LuCHZQY23}
\bibfield{author}{\bibinfo{person}{Zhichao Lu}, \bibinfo{person}{Ran Cheng}, \bibinfo{person}{Shihua Huang}, \bibinfo{person}{Haoming Zhang}, \bibinfo{person}{Changxiao Qiu}, {and} \bibinfo{person}{Fan Yang}.} \bibinfo{year}{2023}\natexlab{}.
\newblock \showarticletitle{Surrogate-Assisted Multiobjective Neural Architecture Search for Real-Time Semantic Segmentation}.
\newblock \bibinfo{journal}{\emph{{IEEE} Transactions on Artificial Intelligence}} \bibinfo{volume}{4}, \bibinfo{number}{6} (\bibinfo{year}{2023}), \bibinfo{pages}{1602--1615}.
\newblock


\bibitem[Lu et~al\mbox{.}(2024)]%
        {10004638}
\bibfield{author}{\bibinfo{person}{Zhichao Lu}, \bibinfo{person}{Ran Cheng}, \bibinfo{person}{Yaochu Jin}, \bibinfo{person}{Kay~Chen Tan}, {and} \bibinfo{person}{Kalyanmoy Deb}.} \bibinfo{year}{2024}\natexlab{}.
\newblock \showarticletitle{Neural Architecture Search as Multiobjective Optimization Benchmarks: Problem Formulation and Performance Assessment}.
\newblock \bibinfo{journal}{\emph{IEEE Transactions on Evolutionary Computation}} \bibinfo{volume}{28}, \bibinfo{number}{2} (\bibinfo{year}{2024}), \bibinfo{pages}{323--337}.
\newblock


\bibitem[Tian et~al\mbox{.}(2017)]%
        {PlatEMO}
\bibfield{author}{\bibinfo{person}{Ye Tian}, \bibinfo{person}{Ran Cheng}, \bibinfo{person}{Xingyi Zhang}, {and} \bibinfo{person}{Yaochu Jin}.} \bibinfo{year}{2017}\natexlab{}.
\newblock \showarticletitle{{PlatEMO}: A {MATLAB} platform for evolutionary multi-objective optimization}.
\newblock \bibinfo{journal}{\emph{IEEE Computational Intelligence Magazine}} \bibinfo{volume}{12}, \bibinfo{number}{4} (\bibinfo{year}{2017}), \bibinfo{pages}{73--87}.
\newblock


\bibitem[Zhang and Li(2007)]%
        {DBLP:journals/tec/ZhangL07}
\bibfield{author}{\bibinfo{person}{Qingfu Zhang} {and} \bibinfo{person}{Hui Li}.} \bibinfo{year}{2007}\natexlab{}.
\newblock \showarticletitle{{MOEA/D:} {A} Multiobjective Evolutionary Algorithm Based on Decomposition}.
\newblock \bibinfo{journal}{\emph{{IEEE} Transactions on Evolutionary Computation}} \bibinfo{volume}{11}, \bibinfo{number}{6} (\bibinfo{year}{2007}), \bibinfo{pages}{712--731}.
\newblock


\bibitem[Zitzler and K{\"{u}}nzli(2004)]%
        {DBLP:conf/ppsn/ZitzlerK04}
\bibfield{author}{\bibinfo{person}{Eckart Zitzler} {and} \bibinfo{person}{Simon K{\"{u}}nzli}.} \bibinfo{year}{2004}\natexlab{}.
\newblock \showarticletitle{Indicator-Based Selection in Multiobjective Search}. In \bibinfo{booktitle}{\emph{Parallel Problem Solving from Nature - {PPSN} VIII, 8th International Conference, Birmingham, UK, September 18-22, 2004, Proceedings}}. \bibinfo{pages}{832--842}.
\newblock


\end{thebibliography}


\appendix
\newpage
\onecolumn{}

\section{Experimental Setup}
\label{appendix:Experimental Setup}
The experiments were conducted on EvoX \cite{evox} for Python, and PlatEMO \cite{PlatEMO} for MATLAB.
\begin{table}[ht]
  \centering
    \caption{Definition of objectives in CitySeg/MOP test suite.}
  \begin{tabular}{ccc}  
  \toprule
  Objectives & Definition \\
  \midrule
    \rule{0pt}{12pt} $f^e$ & prediction error \\
    \rule{0pt}{12pt} $f^{h_1}_1$ & $h_1$'s inference latency \\
    \rule{0pt}{12pt} $f^{h_2}_1$ & $h_2$'s inference latency \\
    \rule{0pt}{12pt} $f^{h_1}_2$ & $h_1$'s inference energy consumption \\
    \rule{0pt}{12pt} $f^{h_2}_2$ & $h_2$'s inference energy consumption \\
    \rule{0pt}{12pt} $f^c_1$ & \# of floating point operations \\
    \rule{0pt}{12pt} $f^c_2$ & \# of parameters/weights \\
  \bottomrule
  \end{tabular}
  \label{table:definition of objectives}
\end{table}

\begin{table}[ht]
  \centering
    \caption{Summary of experimental setup.}
  \begin{tabular}{ccc}  
  \toprule
  Parameter & Setting \\
  \midrule
    \# of runs & 31 \\
    \# of evaluations & 10,000 \\
    $D$, \# of decision variables &  32 \\
    
  \bottomrule
  \end{tabular}
  \label{table:experimental setup}
\end{table}

\begin{table}[ht]
  \centering
    \caption{Population size settings. The size of the population $N$ is set corresponding to the number of objectives $M$.}
  \begin{tabular}{ccc}  
  \toprule
  $M$ & $(H_1, H_2)$ & $N$ \\
  \midrule
    2 & (99, 0) & 100 \\
    3 & (13, 0) & 105 \\
    4 & (7, 0) & 120 \\
    5 & (5, 0) & 126 \\
    6 & (4, 1) & 132 \\
    7 & (3, 2) & 217 \\
  \bottomrule
  \end{tabular}
  \label{table:population size}
\end{table}

\begin{table*}[h]
  \caption{The reference point used for calculating the hypervolume of CitySeg/MOP.} 
  \label{tab: Reference point of HV metric.}
  \begin{tabular}{ccc}
  \toprule
  Problem  &Reference Point \\
  \midrule
  CitySeg/MOP1  &[0.0, 1.9741] \\
  CitySeg/MOP2  &[0.0, 1.9741, 3.3107e8]    \\
  CitySeg/MOP3  &[0.0, 1.9741, 1.3251e5]     \\
  CitySeg/MOP4  &[0.0, 1.9741, 678.0692, 3.3107e8]    \\
  CitySeg/MOP5  &[0.0, 1.9741, 678.0692, 3.3107e8, 1.3251e5]     \\
  CitySeg/MOP6  &[0.0, 58.7465]     \\
  CitySeg/MOP7  &[0.0, 58.7465, 3.3107e8]     \\
  CitySeg/MOP8  &[0.0, 58.7465, 1.3251e5]     \\
  CitySeg/MOP9  &[0.0, 58.7465, 734.3339, 3.3107e8]     \\
  CitySeg/MOP10  &[0.0, 58.7465, 734.3339, 3.3107e8, 1.3251e5]     \\
  CitySeg/MOP11  &[0.0, 1.9741, 58.7465]     \\
  CitySeg/MOP12  &[0.0, 1.9741, 58.7465, 678.0692, 734.3339]     \\
  CitySeg/MOP13  &[0.0, 1.9741, 58.7465, 678.0692, 734.3339, 3.3107e8]     \\
  CitySeg/MOP14  &[0.0, 1.9741, 58.7465, 678.0692, 734.3339, 1.3251e5]     \\
  CitySeg/MOP15  &[0.0, 1.9741, 58.7465, 678.0692, 734.3339, 3.3107e8, 1.3251e5]     \\
\bottomrule
\end{tabular}
\end{table*}

\newpage

\section{Supplementary Results}
\label{appendix:Supplementary Results}

\begin{table}[ht]
  \centering
    \caption{Average time cost (in seconds) of evaluating 100 architectures performance with Python or MATLAB interface. Results are estimated by averaging 31 runs on NVIDIA GeForce RTX 4090 for Python, and one CPU core of a local machine for MATLAB.
}
  \begin{tabular}{ccc}  
  \toprule
  Problem & Python & MATLAB \\
  \midrule
    CitySeg/MOP1 & $1.7776 \pm\ 0.0383$ & $2.5191 \pm 0.0049$ \\
    CitySeg/MOP2 & $1.8190 \pm\ 0.0253$ & $2.5058 \pm 0.0034$ \\
    CitySeg/MOP3 & $1.8336 \pm\ 0.0214$ & $2.5053 \pm 0.0049$ \\
    CitySeg/MOP4 & $1.8774 \pm\ 0.0274$ & $2.4861 \pm 0.0045$ \\
    CitySeg/MOP5 & $1.8752 \pm\ 0.0271$ & $2.3957 \pm 0.0072$ \\
    CitySeg/MOP6 & $1.8445 \pm\ 0.0230$ & $2.3964 \pm 0.0022$ \\
    CitySeg/MOP7 & $1.8719 \pm\ 0.0274$ & $2.3977 \pm 0.0016$ \\
    CitySeg/MOP8 & $1.8563 \pm\ 0.0345$ & $2.3999 \pm 0.0020$ \\
    CitySeg/MOP9 & $1.8563 \pm\ 0.0323$ & $2.3949 \pm 0.0019$ \\
    CitySeg/MOP10 & $1.8473 \pm\ 0.0226$ & $2.3845 \pm 0.0018$ \\
    CitySeg/MOP11 & $1.8357 \pm\ 0.0272$ & $2.4067 \pm 0.0025$ \\
    CitySeg/MOP12 & $1.8362 \pm\ 0.0182$ & $2.3969 \pm 0.0017$ \\
    CitySeg/MOP13 & $1.8343 \pm\ 0.0172$ & $2.3903 \pm 0.0015$ \\
    CitySeg/MOP14 & $1.8340 \pm\ 0.0247$ & $2.3956 \pm 0.0021$ \\
    CitySeg/MOP15 & $2.0100 \pm\ 0.0622$ & $2.3653 \pm 0.0018$ \\
  \bottomrule
  \end{tabular}
  \label{table:time cost}
\end{table}

\begin{figure}
  \centering

  \newcommand{\subfigwidth}{0.24\linewidth} 
  \begin{subfigure}{\subfigwidth}
    \centering
    \includegraphics[width=\linewidth]{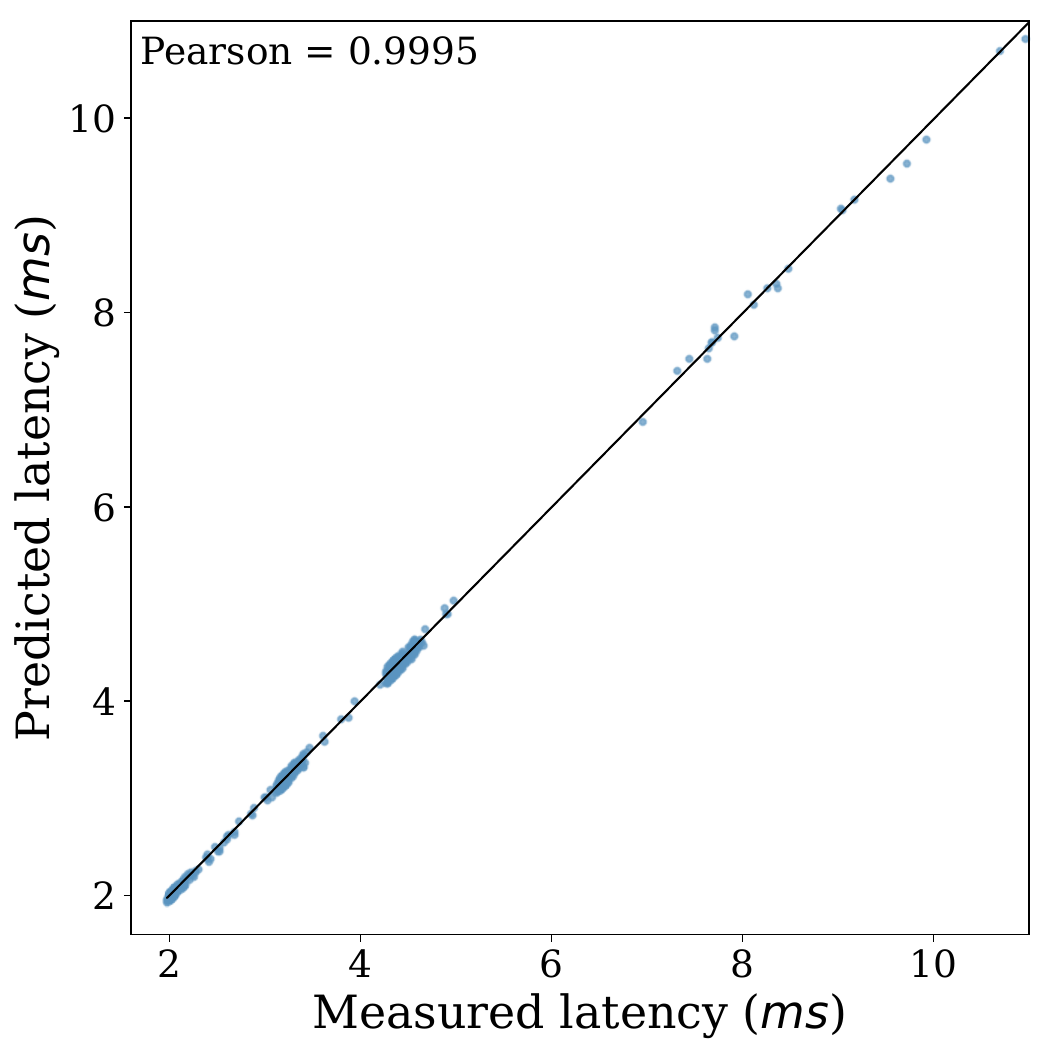}
    \caption{$f^{h_1}_1$}
    \label{fig:h1 latency corrolation}
  \end{subfigure}
  \hfill
  \begin{subfigure}{\subfigwidth}
    \centering
    \includegraphics[width=\linewidth]{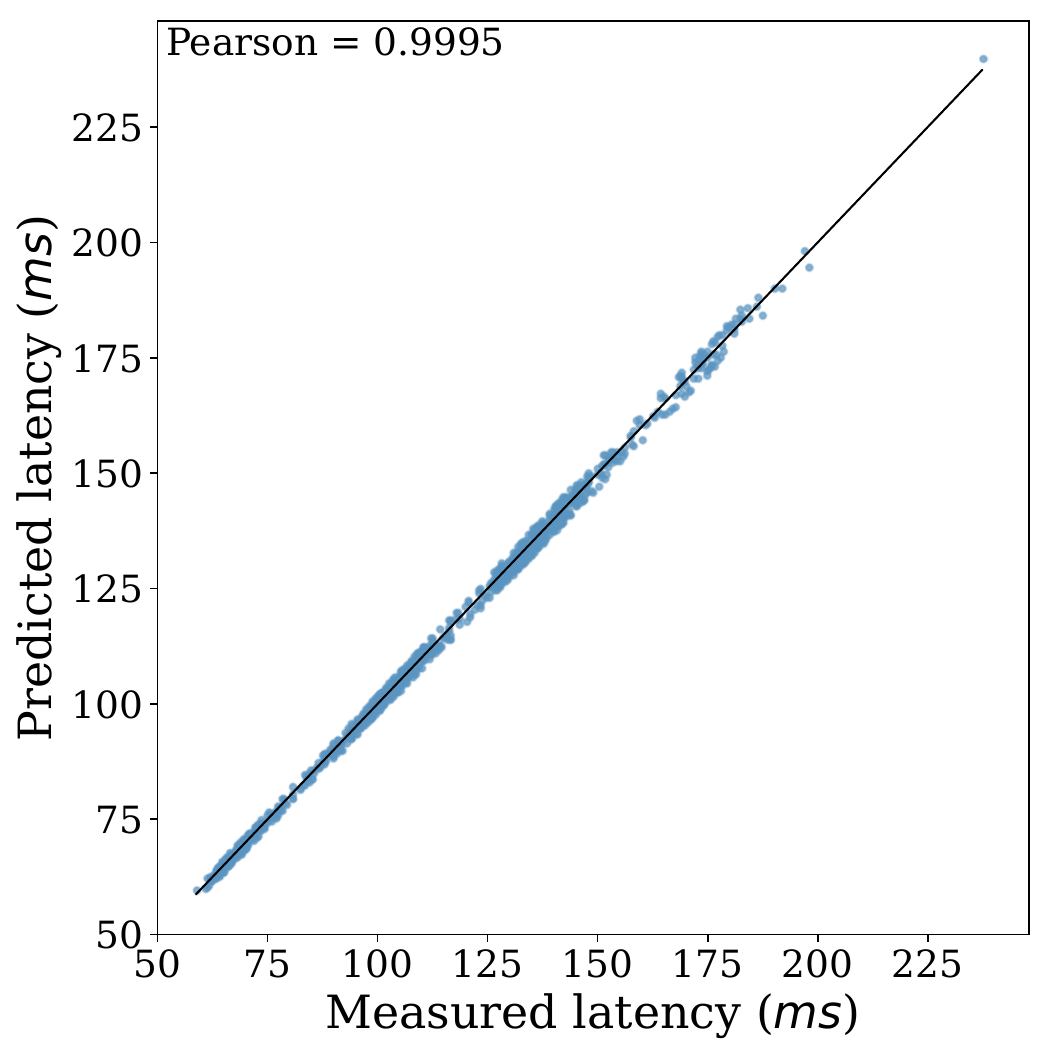}
    \caption{$f^{h_2}_1$}
    \label{fig:h2 latency corrolation}
  \end{subfigure}
  \hfill
  \begin{subfigure}{\subfigwidth}
    \centering
    \includegraphics[width=\linewidth]{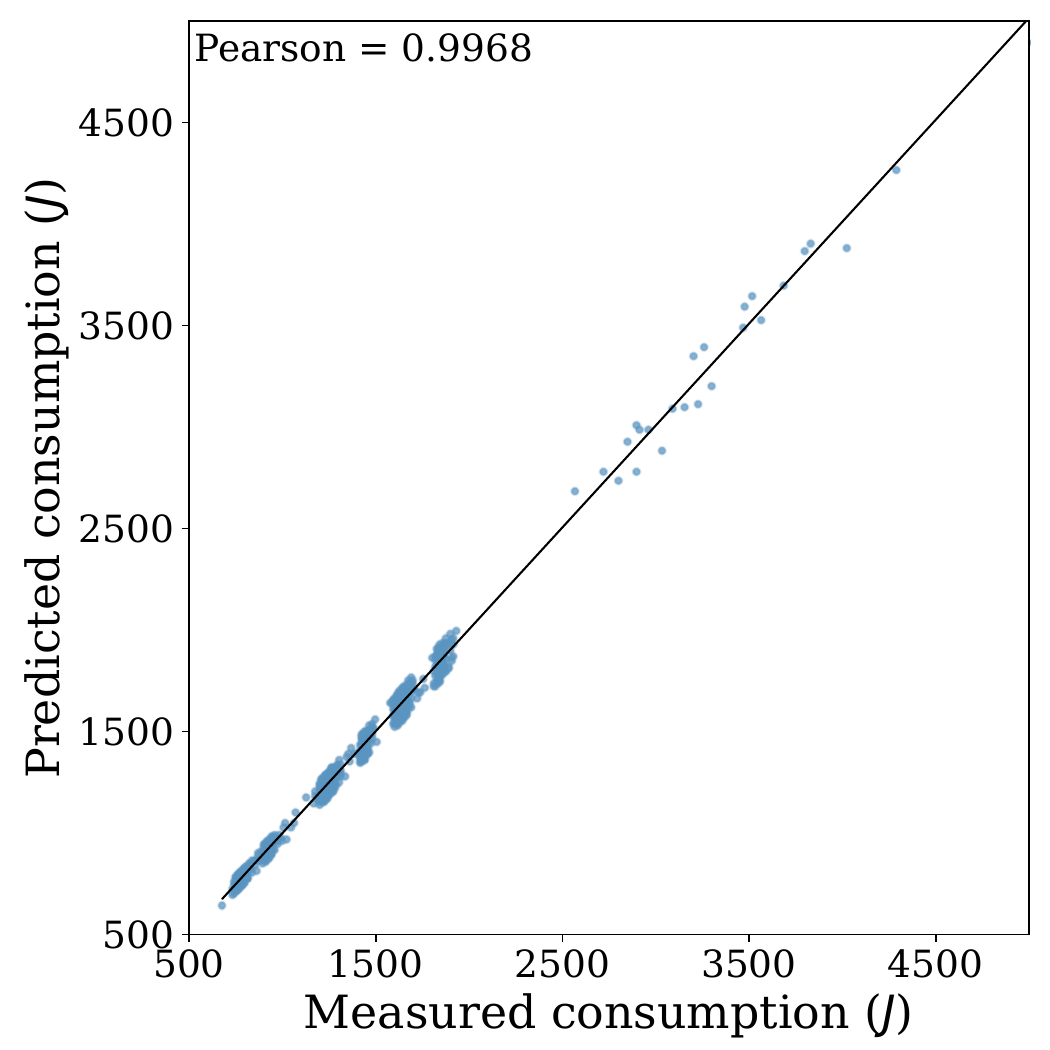}
    \caption{$f^{h_1}_2$}
    \label{fig:h1 consumption corrolation}
  \end{subfigure}
  \hfill
  \begin{subfigure}{\subfigwidth}
    \centering
    \includegraphics[width=\linewidth]{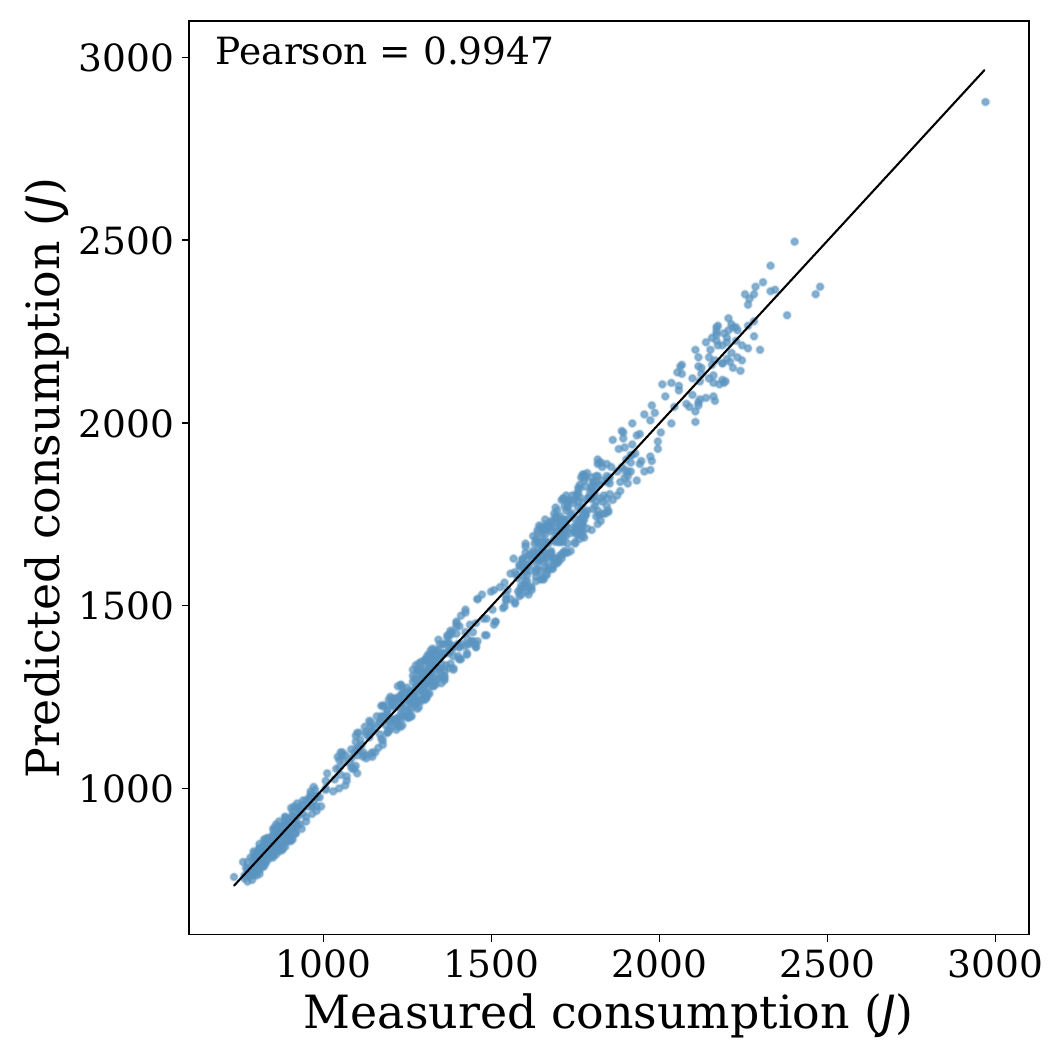}
    \caption{$f^{h_2}_2$}
    \label{fig:h2 consumption corrolation}
  \end{subfigure}
  \caption{Correlations between the measured and the predicted metrics. The measured and the predicted metrics are obtained from measurements and the benchmark test suite correspondingly.
  } 
  \label{fig:latency and consumption correlation}
\end{figure}

\begin{figure}
  \centering

  \newcommand{\subfigwidth}{0.45\linewidth}

    \begin{subfigure}{\subfigwidth}
    \centering
    \includegraphics[width=\linewidth]{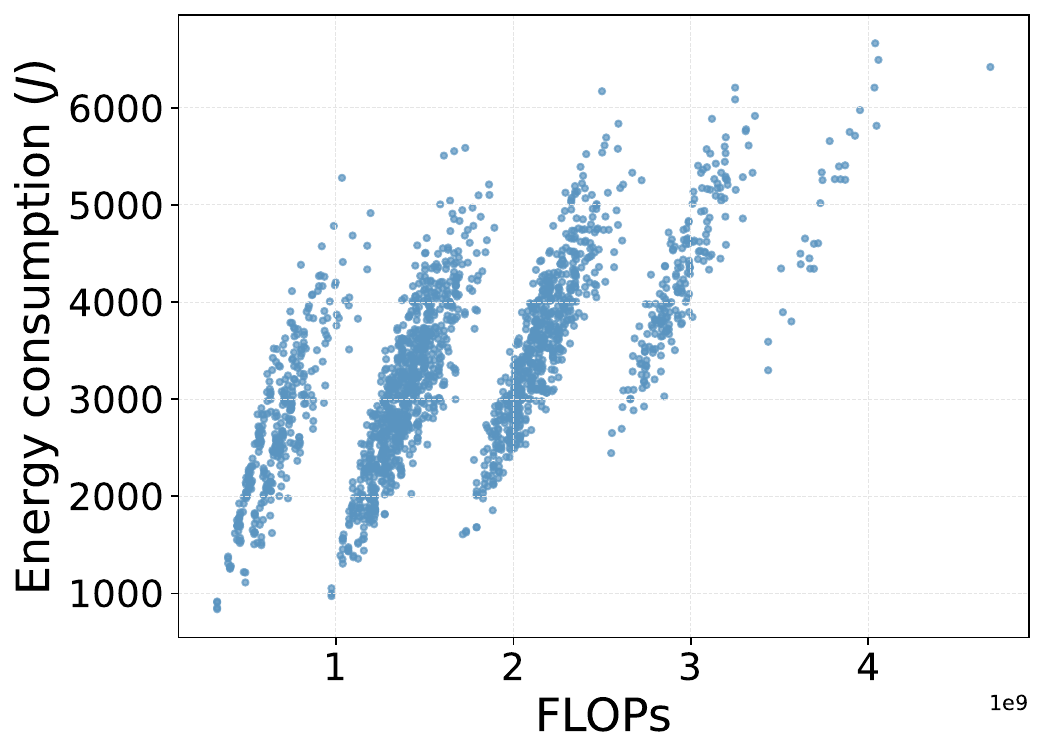}
    \caption{$f^c_1$ and $f^h_2$'s correlation.}
    \label{fig:$f^c_1$ and $f^h_2$'s correlation.}
  \end{subfigure}
  \hfill
  \begin{subfigure}{\subfigwidth}
    \centering
    \includegraphics[width=\linewidth]{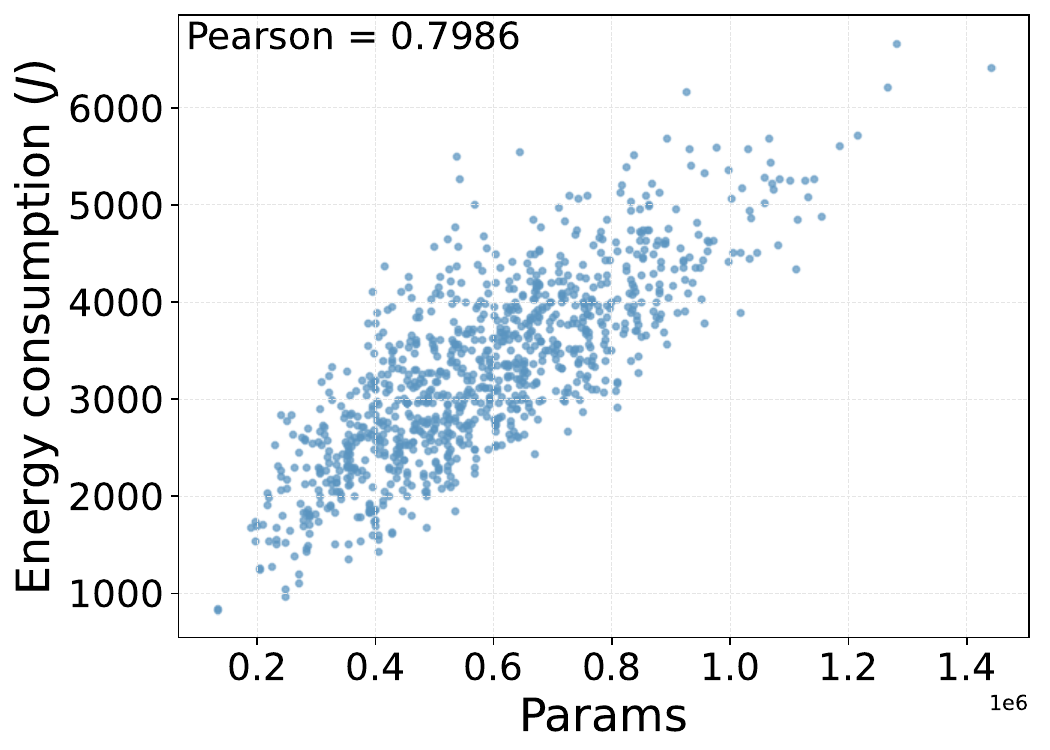}
    \caption{$f^c_2$ and $f^h_2$'s correlation.}
    \label{fig:$f^c_2$ and $f^h_2$'s correlation.}
  \end{subfigure}
  \caption{Correlation between $f^h_2$ and $f^c_1 / f^c_2$ under randomly sampled architectures.}
  \label{fig:architectures' correlation between consumption and flops and params}
\end{figure}

\begin{table*}
  \caption{Statistical results (mean value and standard deviation) of the hypervolume values of the test instances on CitySeg/MOP test suite. The best results obtained from each instance are highlighted in bold. The superscripts $+$, $-$, and $\approx$ denote the best, the worst, and identically distributed result obtained by the six algorithms. Whether the two sets of data are identically distributed or not is obtained from the Rank Sum Test.} 
  \label{tab: problem result of CitySeg/MOP}
  \begin{tabular*}{\hsize}{@{}@{\extracolsep{\fill}}ccccccc@{}}
  \toprule
  Problem  &NSGA-II  &NSGA-III &MOEA/D &RVEA &IBEA &HypE\\
  \midrule
CitySeg/MOP1  &$\mathbf{0.9001 (0.0045)}^\approx$ &$\mathbf{0.8983 (0.0071)}^\approx$ &${0.8423 (0.0505)}^-$ &${0.8683 (0.0229)}^-$ &${0.8990 (0.0043)}^-$ &${0.8967 (0.0111)}^-$ \\

CitySeg/MOP2  &$\mathbf{0.7995 (0.0016)}^\approx$ &$\mathbf{0.7991 (0.0020)}^\approx$ &${0.7492 (0.0415)}^-$ &${0.6976 (0.0705)}^-$ &${0.7739 (0.0174)}^-$ &${0.7949 (0.0050)}^-$ \\

CitySeg/MOP3  &$\mathbf{0.8238 (0.0022)}^\approx$ &$\mathbf{0.8229 (0.0036)}^\approx$ &${0.7582 (0.0333)}^-$ &${0.7813 (0.0320)}^-$ &${0.7830 (0.0076)}^-$ &${0.8134 (0.0181)}^-$ \\

CitySeg/MOP4  &$\mathbf{0.6979 (0.0011)}^\approx$ &$\mathbf{0.6983 (0.0007)}^\approx$ &${0.5474 (0.0854)}^-$ &${0.5951 (0.0695)}^-$ &${0.6006 (0.0503)}^-$ &$\mathbf{0.6979 (0.0014)}^\approx$ \\

CitySeg/MOP5  &$\mathbf{0.6562 (0.0009)}^\approx$ &$\mathbf{0.6565 (0.0006)}^\approx$ &${0.5576 (0.0451)}^-$ &${0.5315 (0.1117)}^-$ &${0.5612 (0.0420)}^-$ &$\mathbf{0.6563 (0.0006)}^\approx$ \\

CitySeg/MOP6  &$\mathbf{0.7719 (0.0001)}^\approx$ &$\mathbf{0.7718 (0.0003)}^\approx$ &${0.7107 (0.0541)}^-$ &${0.7367 (0.0227)}^-$ &$\mathbf{0.7692 (0.0091)}^\approx$ &$\mathbf{0.7706 (0.0073)}^\approx$ \\

CitySeg/MOP7  &$\mathbf{0.7311 (0.0003)}^\approx$ &$\mathbf{0.7312 (0.0002)}^\approx$ &${0.6745 (0.0488)}^-$ &${0.6810 (0.0297)}^-$ &${0.7090 (0.0339)}^-$ &$\mathbf{0.7303 (0.0030)}^\approx$ \\

CitySeg/MOP8  &$\mathbf{0.7324 (0.0002)}^\approx$ &$\mathbf{0.7324 (0.0002)}^\approx$ &${0.6843 (0.0342)}^-$ &${0.6889 (0.0314)}^-$ &${0.7225 (0.0223)}^-$ &${0.7284 (0.0128)}^-$ \\

CitySeg/MOP9  &$\mathbf{0.5767 (0.0004)}^\approx$ &$\mathbf{0.5768 (0.0003)}^\approx$ &${0.4804 (0.0373)}^-$ &${0.5405 (0.0354)}^-$ &${0.4923 (0.0409)}^-$ &$\mathbf{0.5768 (0.0004)}^\approx$ \\

CitySeg/MOP10  &$\mathbf{0.5473 (0.0003)}^\approx$ &$\mathbf{0.5473 (0.0003)}^\approx$ &${0.4606 (0.0386)}^-$ &${0.5138 (0.0325)}^-$ &${0.4726 (0.0315)}^-$ &$\mathbf{0.5472 (0.0004)}^\approx$ \\

CitySeg/MOP11  &$\mathbf{0.6884 (0.0010)}^\approx$ &$\mathbf{0.6884 (0.0009)}^\approx$ &${0.5699 (0.0351)}^-$ &${0.6620 (0.0359)}^-$ &${0.6634 (0.0213)}^-$ &${0.6840 (0.0097)}^-$ \\

CitySeg/MOP12  &${0.4688 (0.0019)}^-$ &$\mathbf{0.4705 (0.0012)}^+$ &${0.3558 (0.0651)}^-$ &${0.4358 (0.0294)}^-$ &${0.3889 (0.0283)}^-$ &${0.4527 (0.0172)}^-$ \\

CitySeg/MOP13  &${0.4184 (0.0051)}^-$ &$\mathbf{0.4228 (0.0006)}^+$ &${0.3459 (0.0212)}^-$ &${0.3772 (0.0217)}^-$ &${0.3503 (0.0349)}^-$ &${0.4150 (0.0093)}^-$ \\

CitySeg/MOP14  &${0.4299 (0.0028)}^-$ &$\mathbf{0.4339 (0.0007)}^+$ &${0.3406 (0.0200)}^-$ &${0.4118 (0.0173)}^-$ &${0.3569 (0.0240)}^-$ &${0.4214 (0.0157)}^-$ \\

CitySeg/MOP15  &${0.3971 (0.0018)}^-$ &$\mathbf{0.3980 (0.0025)}^+$ &${0.3272 (0.0265)}^-$ &${0.3003 (0.0718)}^-$ &${0.3376 (0.0192)}^-$ &${0.3956 (0.0045)}^-$ \\

\bottomrule
\end{tabular*}
\end{table*}

\begin{figure*}
  \centering
  \newcommand{\subfigwidth}{0.32\linewidth} 
  \begin{subfigure}{\subfigwidth}
    \centering
    \includegraphics[width=\linewidth]{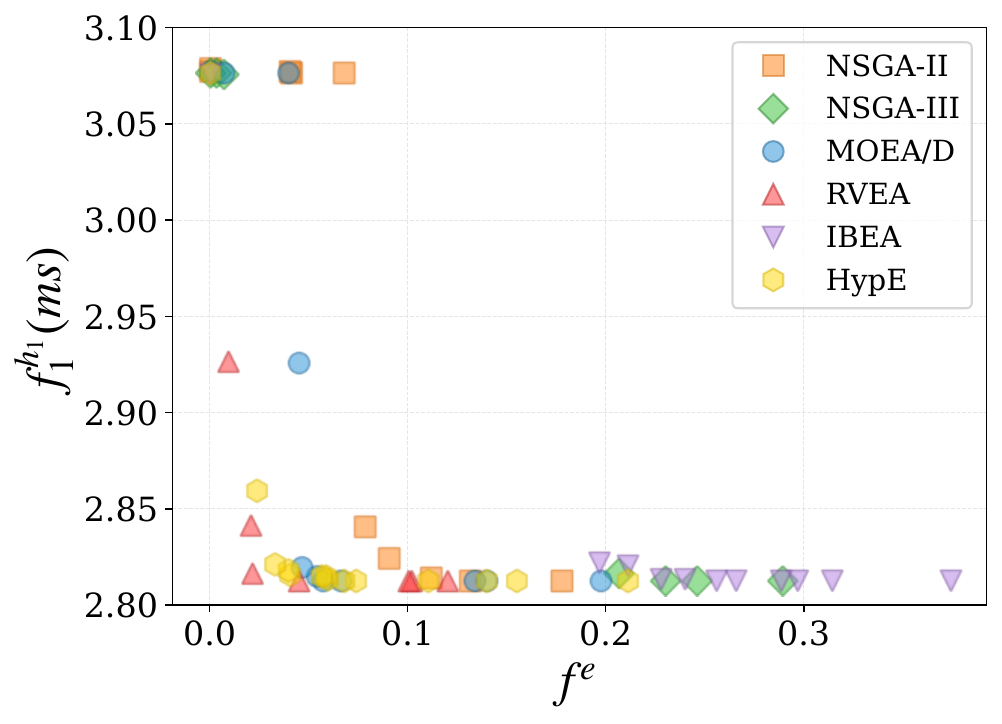}
    \caption{Result of CitySeg/MOP1.}
    \label{fig:citymop1_all}
  \end{subfigure}
  \hfill
  \begin{subfigure}{\subfigwidth}
    \centering
    \includegraphics[width=\linewidth]{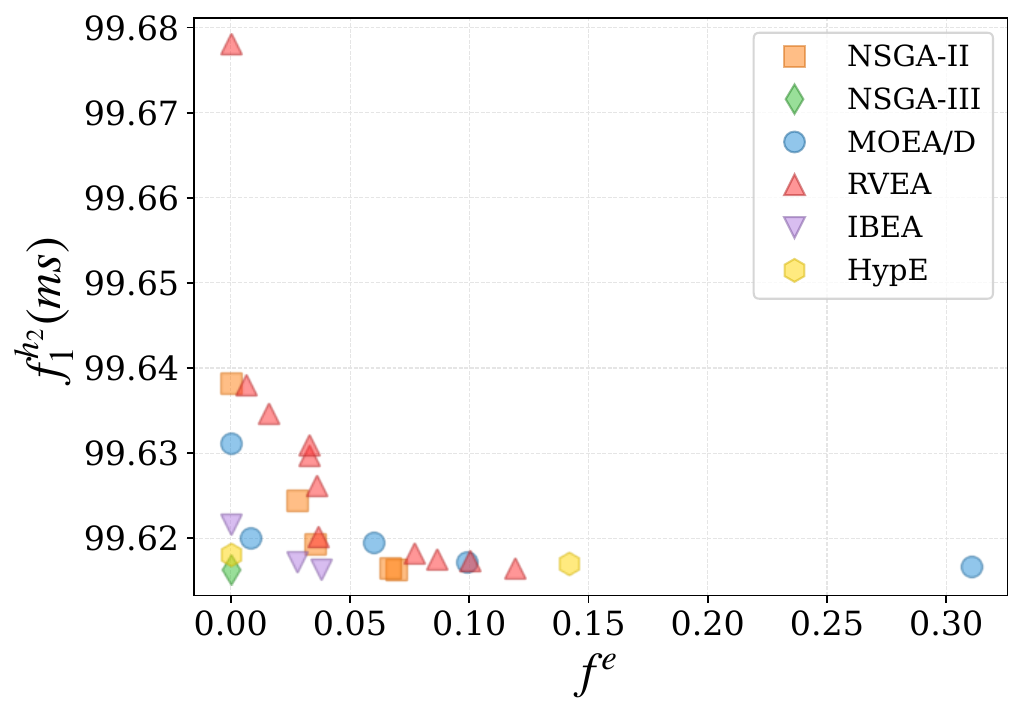}
    \caption{Result of CitySeg/MOP6.}
    \label{fig:citymop6_all}
  \end{subfigure}
  \hfill
  \begin{subfigure}{\subfigwidth}
    \centering
    \includegraphics[width=\linewidth]{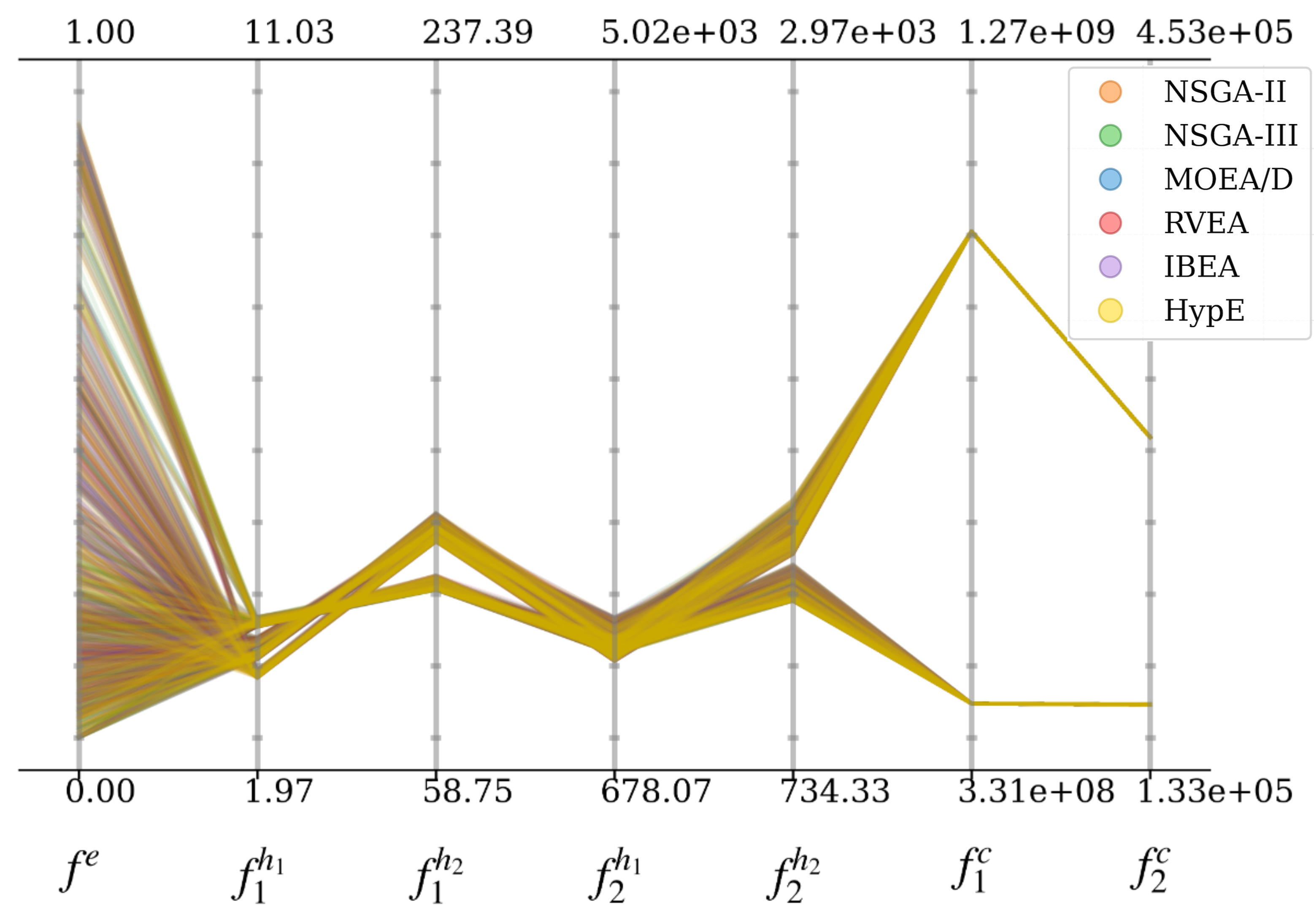}
    \caption{Result of CitySeg/MOP15.}
    \label{fig:citymop15_all}
  \end{subfigure}
  \caption{Non-dominated solutions obtained by each algorithm on CitySeg/MOP1, CitySeg/MOP6 and CitySeg/MOP15. 
  } 
  \label{fig:citymop problems}
\end{figure*}

\end{document}